\documentclass[runningheads]{llncs}
\usepackage{graphicx}
\usepackage{tikz}
\usepackage{comment}
\usepackage{amsmath,amssymb} % define this before the line numbering.
\usepackage{color}
\usepackage{xspace}

\makeatletter

\usepackage{pgfplots} % For plots
\usepackage{pgfplotstable} % For plots
\pgfplotsset{compat=newest} % For plots
\usetikzlibrary{plotmarks} % For plots
\usetikzlibrary{colorbrewer} % Nice colors

\usepackage{algorithm}
\usepackage[noend]{algorithmic}
\usepackage{graphicx, amsmath, amssymb, caption, subcaption, multirow, overpic, textpos}

\DeclareRobustCommand\onedot{\futurelet\@let@token\@onedot}
\def\@onedot{\ifx\@let@token.\else.\null\fi\xspace}

\def\eg{\emph{e.g}\onedot} 
\def\ie{\emph{i.e}\onedot}

\def\wrt{w.r.t\onedot}

\newcommand{\figref}[1]{Fig\onedot~\ref{#1}}
\newcommand{\equref}[1]{Eq\onedot~\eqref{#1}}

\newcommand{\tabref}[1]{Tab\onedot~\ref{#1}}

\newlength\secmargin
\newlength\paramargin
\newlength\abovetabcapmargin
\newlength\belowtabcapmargin
\newlength\abovefigcapmargin
\newlength\belowfigcapmargin

\usepackage{pifont}

\usepackage[pagebackref=true,breaklinks=true,colorlinks,bookmarks=false]{hyperref}
\usepackage{xcolor}
\usepackage{colortbl}

\definecolor{baselinecolor}{gray}{.9}
\newcommand{\baseline}[1]{\cellcolor{baselinecolor}{#1}}

\newlength\savewidth\newcommand\shline{\noalign{\global\savewidth\arrayrulewidth
  \global\arrayrulewidth 1pt}\hline\noalign{\global\arrayrulewidth\savewidth}}
\newcommand{\tablestyle}[2]{\setlength{\tabcolsep}{#1}\renewcommand{\arraystretch}{#2}\centering\footnotesize}
\usepackage{xcolor}

\begin{document}
\pagestyle{headings}
\mainmatter

\title{kMaX-DeepLab: k-Means Mask Transformer}

% INITIAL SUBMISSION 
\begin{comment}
\titlerunning{ECCV-22 submission ID \ECCVSubNumber} 
\authorrunning{ECCV-22 submission ID \ECCVSubNumber} 
\author{Anonymous ECCV submission}
\institute{Paper ID \ECCVSubNumber}
\end{comment}
%******************

% CAMERA READY SUBMISSION
%\begin{comment}
\titlerunning{$k$-means Mask Transformer}
% If the paper title is too long for the running head, you can set
% an abbreviated paper title here
%
\author{Qihang Yu\inst{1} \and
Huiyu Wang\inst{1} \and
Siyuan Qiao\inst{2} \and
Maxwell Collins\inst{2} \and
Yukun Zhu\inst{2} \and
Hartwig Adam\inst{2} \and
Alan Yuille\inst{1} \and
Liang-Chieh Chen\inst{2}}
\authorrunning{Q. Yu et al.}
% First names are abbreviated in the running head.
% If there are more than two authors, 'et al.' is used.
%
\institute{Johns Hopkins University \and
Google Research}
%\end{comment}
%******************
\maketitle

\begin{abstract}
The rise of transformers in vision tasks not only advances network backbone designs, but also starts a brand-new page to achieve end-to-end image recognition (\eg, object detection and panoptic segmentation).
Originated from Natural Language Processing (NLP), transformer architectures, consisting of self-attention and cross-attention, effectively learn long-range interactions between elements in a sequence.
However, we observe that most existing transformer-based vision models simply borrow the idea from NLP, neglecting the crucial difference between languages and images, particularly the extremely large sequence length of spatially flattened pixel features.
This subsequently impedes the learning in cross-attention between pixel features and object queries.
In this paper, we rethink the relationship between pixels and object queries, and propose to reformulate the cross-attention learning as a clustering process.
Inspired by the traditional $k$-means clustering algorithm, we develop a $\mathbf{k}$-means \textbf{Ma}sk \textbf{X}former ($k$MaX-DeepLab) for  segmentation tasks, which not only improves the state-of-the-art, but also enjoys a simple and elegant design.
As a result, our $k$MaX-DeepLab achieves a new state-of-the-art performance on COCO \textit{val} set with 58.0\% PQ, Cityscapes \textit{val} set with 68.4\% PQ, 44.0\% AP, and 83.5\% mIoU, and ADE20K \textit{val} set with 50.9\% PQ and 55.2\% mIoU without test-time augmentation or external dataset. We hope our work can shed some light on designing transformers tailored for vision tasks. TensorFlow code and models are available at \url{https://github.com/google-research/deeplab2}. A PyTorch re-implementation is also available at \url{https://github.com/bytedance/kmax-deeplab}. \let\thefootnote\relax\footnote{$^\ast$Work done during an internship at Google.}
\end{abstract}
\section{Introduction}
Transformers~\cite{vaswani2017attention} are receiving a growing attention in the computer vision community.
On the one hand, the transformer encoder, with multi-head self-attention as the central component, demonstrates a great potential for building powerful network architectures in various visual recognition tasks~\cite{wang2020axial,dosovitskiy2020image,liu2021swin}.
On the other hand, the transformer decoder, with multi-head cross-attention at its core, provides a brand-new approach to tackling complex visual recognition problems in an end-to-end manner, dispensing with hand-designed heuristics.

Recently, the pioneering work DETR~\cite{carion2020end} introduces the first end-to-end object detection system with transformers.
In this framework, the pixel features are firstly extracted by a convolutional neural network~\cite{lecun1998gradient}, followed by the deployment of several transformer encoders for feature enhancement to capture long-range interactions between pixels.
Afterwards, a set of learnable positional embeddings, named object queries, is responsible for interacting with pixel features and aggregating information through several interleaved cross-attention and self-attention modules.
In the end, the object queries, decoded by a Feed-Forward Network (FFN), directly correspond to the final bounding box predictions.
Along the same direction, MaX-DeepLab~\cite{wang2021max} proves the success of transformers in the challenging panoptic segmentation task~\cite{kirillov2018panoptic}, where the prior arts~\cite{kirillov2019panoptic,xiong2019upsnet,cheng2019panopticworkshop} usually adopt complicated pipelines involving hand-designed heuristics. The essence of this framework lies in converting the object queries to mask embedding vectors~\cite{jia2016dynamic,tian2020conditional,wang2020solov2}, which are employed to yield a set of mask predictions by multiplying with the pixel features.

The end-to-end transformer-based frameworks have been successfully applied to multiple computer vision tasks with the help of transformer decoders, especially the cross-attention modules. However, the working mechanism behind the scenes remains unclear. The cross-attention, which arises from the Natural Language Processing (NLP) community, is originally designed for language problems, such as neural machine translation~\cite{sutskever2014sequence,bahdanau2014neural}, where both the input sequence and output sequence share a similar short length. This implicit assumption becomes problematic when it comes to certain vision problems, where the cross-attention is performed between object queries and spatially flattened pixel features with an exorbitantly large length.
Concretely, usually a small number of object queries is employed (\eg, 128 queries), while the input images can contain thousands of pixels for the vision tasks of detection and segmentation. Each object query needs to learn to highlight the most distinguishable features among the abundant pixels in the cross-attention learning process, which subsequently leads to slow training convergence and thus inferior performance~\cite{zhu2020deformable,gao2021fast}.

In this work, we make a crucial observation that the cross-attention scheme actually bears a strong similarity to the traditional $k$-means clustering~\cite{lloyd1982least} by regarding the object queries as cluster centers with learnable embedding vectors.
Our examination of the similarity inspires us to propose the novel $\mathbf{k}$-means \textbf{Ma}sk \textbf{X}former ($k$MaX-DeepLab), which rethinks the relationship between pixel features and object queries, and redesigns the cross-attention from the perspective of $k$-means clustering.
Specifically, when updating the cluster centers (\ie, object queries), our $k$MaX-DeepLab performs a different operation.
Instead of performing \textit{softmax} on the large spatial dimension (image height times  width) as in the original Mask Transformer's cross-attention~\cite{wang2021max}, our $k$MaX-DeepLab performs \textit{argmax} along the cluster center dimension, similar to the $k$-means pixel-cluster assignment step (with a \textit{hard} assignment). 
We then update cluster centers by aggregating the pixel features based on the pixel-cluster assignment (computed by their feature affinity), similar to the $k$-means center-update step.
In spite of being conceptually simple, the modification has a striking impact: on COCO \textit{val} set~\cite{lin2014microsoft}, using the standard ResNet-50~\cite{he2016deep} as backbone, our $k$MaX-DeepLab demonstrates a significant improvement of \textbf{5.2\%} PQ over the original cross-attention scheme at a negligible cost of extra parameters and FLOPs.
When comparing to state-of-the-art methods, our $k$MaX-DeepLab with the simple ResNet-50 backbone already outperforms MaX-DeepLab~\cite{wang2021max} with MaX-L~\cite{wang2021max} backbone by \textbf{1.9\%} PQ, while requiring \textbf{7.9} and \textbf{22.0} times fewer parameters and FLOPs, respectively. Our $k$MaX-DeepLab with ResNet-50 also outperforms MaskFormer~\cite{cheng2021per} with the strong ImageNet-22K pretrained Swin-L~\cite{liu2021swin} backbone, and runs \textbf{4.4} times faster.
Using the modern ConvNeXt-L~\cite{liu2022convnet} as backbone, our $k$MaX-DeepLab further sets a new state-of-the-art performance on the COCO \textit{val} set~\cite{lin2014microsoft} with 58.0\% PQ. It also outperforms other state-of-the-art methods on the Cityscapes \textit{val} set~\cite{Cordts2016Cityscapes}, achieving 68.4\% PQ, 83.5\% mIoU, 44.0\% AP, without using any test-time augmentation or extra dataset pretraining~\cite{lin2014microsoft,neuhold2017mapillary}.
Finally, $k$MaX-DeepLab also advances the new state-of-the-art performance on ADE20K~\cite{zhou2017scene} with 50.9\% PQ and 55.2\% mIoU.

\label{sec:intro}
\section{Related Works}
\label{sec:related}

\textbf{Transformers.}\quad
Transformer~\cite{vaswani2017attention} and its variants~\cite{kitaev2020reformer,wang2020linformer,luong2015effective,child2019generating,beltagy2020longformer,zaheer2020big,gupta2020gmat,ainslie2020etc} have advanced the state-of-the-art in natural language processing tasks~\cite{devlin2019bert,shaw2018self,dai2019transformer} by capturing relations across modalities~\cite{bahdanau2014neural} or in a single context~\cite{cheng2016long,vaswani2017attention}. In computer vision, transformer encoders or self-attention modules are either combined with Convolutional Neural Networks (CNNs)~\cite{wang2018non,buades2005non} or used as standalone backbones~\cite{parmar2019stand,hu2019local,wang2020axial,dosovitskiy2020image,liu2021swin}. Both approaches have boosted various vision tasks, such as image classification~\cite{chen20182,bello2019attention,parmar2019stand,hu2019local,li2020neural,wang2020axial,dosovitskiy2020image,liu2021swin,yu2021glance,yang2022lite}, image generation~\cite{parmar2018image,ho2019axial}, object detection~\cite{wang2018non,shen2021efficient,parmar2019stand,hu2018relation,carion2020end,zhu2020deformable}, video recognition~\cite{wang2018non,chen20182,arnab2021vivit,fan2021multiscale}, semantic segmentation~\cite{chen2016attention,zhao2018psanet,huang2019ccnet,fu2019dual,zhu2019asymmetric,zhu2019empirical,zheng2021rethinking,xie2021segformer,chen2021transunet}, and panoptic segmentation~\cite{wang2020axial}.

\textbf{Mask transformers for segmentation.}\quad
Besides the usage as backbones, transformers are also adopted as task decoders for image segmentation. MaX-DeepLab~\cite{wang2021max} proposed \textbf{Ma}sk \textbf{X}formers (MaX) for end-to-end panoptic segmentation. Mask transformers predict class-labeled object masks and are trained by Hungarian matching the predicted masks with ground truth masks.
The essential component of mask transformers is the conversion of object queries to mask embedding vectors~\cite{jia2016dynamic,tian2020conditional,wang2020solov2}, which are employed to generate predicted masks.
Both Segmenter~\cite{strudel2021segmenter} and MaskFormer~\cite{cheng2021per} applied mask transformers to semantic segmentation.
K-Net~\cite{zhang2021k} proposed dynamic kernels for generating the masks. CMT-DeepLab~\cite{yu2022cmt} proposed to improve the cross-attention with an additional clustering update term. Panoptic Segformer~\cite{li2021panoptic} strengthened mask transformer with deformable attention~\cite{zhu2020deformable}, while Mask2Former~\cite{cheng2021masked} further boosted the performance with masked cross-attention along with a series of technical improvements including cascaded transformer decoder, deformable attention~\cite{zhu2020deformable}, uncertainty-based pointly supervision~\cite{kirillov2020pointrend}, \emph{etc}.
These mask transformer methods generally outperform box-based methods~\cite{kirillov2019panoptic} that decompose panoptic segmentation into multiple surrogate tasks (\eg, predicting masks for each detected object bounding box~\cite{he2017mask}, followed by fusing the instance segments (`thing') and semantic segments (`stuff')~\cite{chen2018deeplabv2} with merging modules~\cite{li2018attention,porzi2019seamless,liu2019e2e,yang2020sognet,xiong2019upsnet,li2020unifying}). Moreover, mask transformers showed great success in the video segmentation problems~\cite{kim2022tubeformer,cheng2021mask2former,li2022video}.

\textbf{Clustering methods for segmentation.}\quad
Traditional image segmentation methods~\cite{lloyd1982least,zhu1996region,achanta2012slic} typically cluster image intensities into a set of masks or superpixels with gradual growing or refinement. However, it is challenging for these traditional methods to capture high-level semantics.
Modern clustering-based methods usually operate on semantic segments~\cite{deeplabv12015,chen2017deeplabv3,deeplabv3plus2018} and group `thing' pixels into instance segments with various representations, such as instance center regression~\cite{kendall2018multi,uhrig2018box2pix,neven2019instance,yang2019deeperlab,cheng2019panoptic,wang2020axial,li2021fully}, Watershed transform \cite{vincent1991watersheds,bai2017deep}, Hough-voting \cite{ballard1981generalizing,leibe2004combined,Wang_2020_CVPR}, or pixel affinity \cite{keuper2015efficient,liu2018affinity,sofiiuk2019adaptis,gao2019ssap,hwang2019segsort}.

Recently, CMT-DeepLab~\cite{yu2022cmt} discussed the similarity between mask transformers and clustering algorithms. However, they only used the clustering update as a complementary term in the cross-attention. In this work, we further discover the underlying similarity between mask transformers and the $k$-means clustering algorithm, resulting in a simple yet effective $k$-means mask transformer.
Finally, we note that several recent works~\cite{locatello2020object,xu2022groupvit,yu2022cmt,zhou2022slot} revisited the relationship between query and key in the attention operation. They applied the cross-attention softmax operation along the query dimension and showed promising results.
%, regardless of the difference in model architectures and target tasks.
\section{Method}
In this section, we first overview the mask-transformer-based segmentation framework presented by MaX-DeepLab~\cite{wang2021max}.
We then revisit the transformer cross-attention~\cite{vaswani2017attention} and the $k$-means clustering algorithm~\cite{lloyd1982least}, and reveal their underlying similarity.
Afterwards, we introduce the proposed $\mathbf{k}$-means \textbf{Ma}sk \textbf{X}former ($k$MaX-DeepLab), which redesigns the cross-attention from a clustering perspective. Even though simple, $k$MaX-DeepLab effectively and significantly improves the segmentation performance.

\subsection{Mask-Transformer-Based Segmentation Framework}

Transformers~\cite{vaswani2017attention} have been effectively deployed to segmentation tasks.
Without loss of generality, we consider panoptic segmentation~\cite{kirillov2018panoptic} in the following problem formulation, which can be easily generalized to other segmentation tasks.

\textbf{Problem statement.}\quad
Panoptic segmentation aims to segment the image $\mathbf{I} \in \mathbb{R}^{H \times W \times 3}$ into a set of non-overlapping masks with associated semantic labels:
\begin{equation}
\{y_i\}_{i=1}^K = \{(m_i, c_i)\}_{i=1}^K \,.
\end{equation}
The $K$ ground truth masks $m_i \in {\{0,1\}}^{H \times W}$ do not overlap with each other, \ie, $\sum_{i=1}^{K} m_i \leq 1^{H \times W}$, and $c_i$ denotes the ground truth class label of mask $m_i$.

Starting from DETR~\cite{carion2020end} and MaX-DeepLab~\cite{wang2021max}, approaches to panoptic segmentation shift to a new end-to-end paradigm, where the prediction directly matches the format of ground-truth with $N$ masks ($N$ is a fixed number and $N\geq K$) and their semantic classes:
\begin{equation}
\{\hat{y_i}\}_{i=1}^N = \{(\hat{m_i}, \hat{p}_{i}(c))\}_{i=1}^N,
\end{equation}
where $\hat{p}_{i}(c)$ denotes the semantic class prediction confidence for the corresponding mask, which includes `thing' classes, `stuff' classes, and the void class $\varnothing$.

The $N$ masks are predicted based on the $N$ object queries, which aggregate information from the pixel features through a transformer decoder, consisting of self-attention and cross-attention modules.

The object queries, updated by multiple transformer decoders, are employed as mask embedding vectors~\cite{jia2016dynamic,tian2020conditional,wang2020solov2}, which will multiply with the pixel features to yield the final prediction $\mathbf{Z} \in \mathbb{R}^{HW \times N}$ that consists of $N$ masks. That is,
\begin{align}
\label{eq:detr_final_output}
\mathbf{Z}& = \operatornamewithlimits{softmax}_{N}(\mathbf{F} \times \mathbf{C}^{\mathrm{T}}),
\end{align}
where $\mathbf{F} \in \mathbb{R}^{HW \times D}$ and $\mathbf{C} \in \mathbb{R}^{N \times D}$ refers to the final pixel features and object queries, respectively. $D$ is the channel dimension of pixel features and object queries. We use underscript $N$ to indicate the axis to perform softmax.

\subsection{Relationship between Cross-Attention and $k$-means Clustering}
Although the transformer-based segmentation frameworks successfully connect object queries and mask predictions in an end-to-end manner, the essential problem becomes how to transform the object queries, starting from learnable embeddings (randomly initialized), into meaningful mask embedding vectors.

{\bf Cross-attention.}\quad
The cross-attention modules are used to aggregate affiliated pixel features to update object queries. Formally, we have
\begin{align}
\label{eq:detr_transformer_update}
\hat{\mathbf{C}}& = \mathbf{C} + \operatornamewithlimits{softmax}_{HW}(\mathbf{Q}^{c} \times (\mathbf{K}^{p})^{\mathrm{T}}) \times \mathbf{V}^{p},
\end{align}
where $\mathbf{C} \in \mathbb{R}^{N \times D}$ refers to $N$ object queries with $D$ channels, and $\hat{\mathbf{C}}$ denotes the updated object queries. We use the underscript $HW$ to represent the axis for softmax on spatial dimension, and superscripts $p$ and $c$ to indicate the feature projected from the pixel features and object queries, respectively. $\mathbf{Q}^c \in \mathbb{R}^{N \times D}, \mathbf{K}^p \in \mathbb{R}^{HW \times D}, \mathbf{V}^p \in \mathbb{R}^{HW \times D} $ stand for the linearly projected features for query, key, and value. For simplicity, we ignore the multi-head mechanism and feed-forward network (FFN) in the equation.

As shown in~\equref{eq:detr_transformer_update}, when updating the object queries, a \textit{softmax} function is applied to the image resolution ($HW$), which is typically in the range of thousands of pixels for the task of segmentation. Given the huge number of pixels, it can take many training iterations to learn the attention map, which starts from a uniform distribution at the beginning (as the queries are randomly initialized). Each object query has a difficult time to identify the most distinguishable features among the abundant pixels in the early stage of training.
This behavior is very different from the application of transformers to natural language processing tasks, \eg, neural machine translation~\cite{sutskever2014sequence,bahdanau2014neural}, where the input and output sequences share a similar short length.
Vision tasks, especially segmentation problems, present another challenge for efficiently learning the cross-attention.

{\bf Discussion.}\quad
Similar to cross-attention, self-attention needs to perform a \textit{softmax} function operated along the image resolution. Therefore, learning the attention map for self-attention may also take many training iterations.
An efficient alternative, such as axial attention~\cite{wang2020axial} or local attention~\cite{liu2021swin} is usually applied on high resolution feature maps, and thus alleviates the problem, while a solution to cross-attention remains an open question for research.

{\bf $k$-means clustering.}\quad
In~\equref{eq:detr_transformer_update}, the cross-attention computes the affinity between object queries and pixels (\ie, $\mathbf{Q}^{c} \times (\mathbf{K}^{p})^{\mathrm{T}}$), which is converted to the attention map through the spatial-wise softmax (operated along the image resolution).
The attention map is then used to retrieve (and weight accordingly) affiliated pixel features to update the object queries. Surprisingly, we observe that the whole process is actually similar to the classic $k$-means clustering algorithm~\cite{lloyd1982least}, which works as follows:
\begin{align}
\label{eq:kmeans_update1}
\mathbf{A}& = \operatornamewithlimits{argmax}_{N}(\mathbf{C} \times \mathbf{P}^{\mathrm{T}}),\\
\label{eq:kmeans_update2}
\hat{\mathbf{C}}& = \mathbf{A} \times \mathbf{P},
\end{align}
where $\mathbf{C}\in \mathbb{R}^{N \times D}$,  $\mathbf{P}\in \mathbb{R}^{HW \times D}$, and $\mathbf{A}\in \mathbb{R}^{N \times HW}$ stand for cluster centers, pixel features, and clustering assignments, respectively.

Comparing~\equref{eq:detr_transformer_update}, ~\equref{eq:kmeans_update1}, and~\equref{eq:kmeans_update2}, we notice that the $k$-means clustering algorithm is parameter-free and thus no linear projection is needed for query, key, and value.
The updates on cluster centers are not in a residual manner.
Most importantly, $k$-means adopts a \textit{cluster-wise argmax} (\ie, argmax operated along the cluster dimension) instead of the spatial-wise softmax when converting the affinity to the attention map (\ie, weights to retrieve and update features).

This observation motivates us to reformulate the cross-attention in vision problems, especially image segmentation.
From a clustering perspective, image segmentation is equivalent to grouping pixels into different clusters, where each cluster corresponds to a predicted mask.
However, the cross-attention mechanism, also attempting to group pixels to different object queries, instead employs a different \textit{spatial-wise softmax} operation from the \textit{cluster-wise argmax} as in $k$-means.
Given the success of $k$-means, we hypothesize that the cluster-wise argmax is a more suitable operation than the spatial-wise softmax regarding pixel clustering, since the cluster-wise argmax performs the hard assignment and efficiently reduces the operation targets from thousands of pixels ($HW$) to just a few cluster centers ($N$), which (we will empirically prove) speeds up the training convergence and leads to a better performance.

\subsection{$k$-means Mask Transformer}
Herein, we first introduce the crucial component of the proposed $k$-means Mask Transformer, \ie, $k$-means cross-attention. We then present its meta architecture and model instantiation.

{\bf $k$-means cross-attention.}\quad
The proposed \textit{$k$-means cross-attention} reformulates the cross-attention in a manner similar to $k$-means clustering:
\begin{align}
\label{eq:kmeans_attention_update}
\hat{\mathbf{C}}& = \mathbf{C} + \operatornamewithlimits{argmax}_{N}(\mathbf{Q}^{c} \times (\mathbf{K}^{p})^{\mathrm{T}}) \times \mathbf{V}^{p}.
\end{align}

\begin{figure}[!t]
    \centering
    \includegraphics[width=0.8\linewidth]{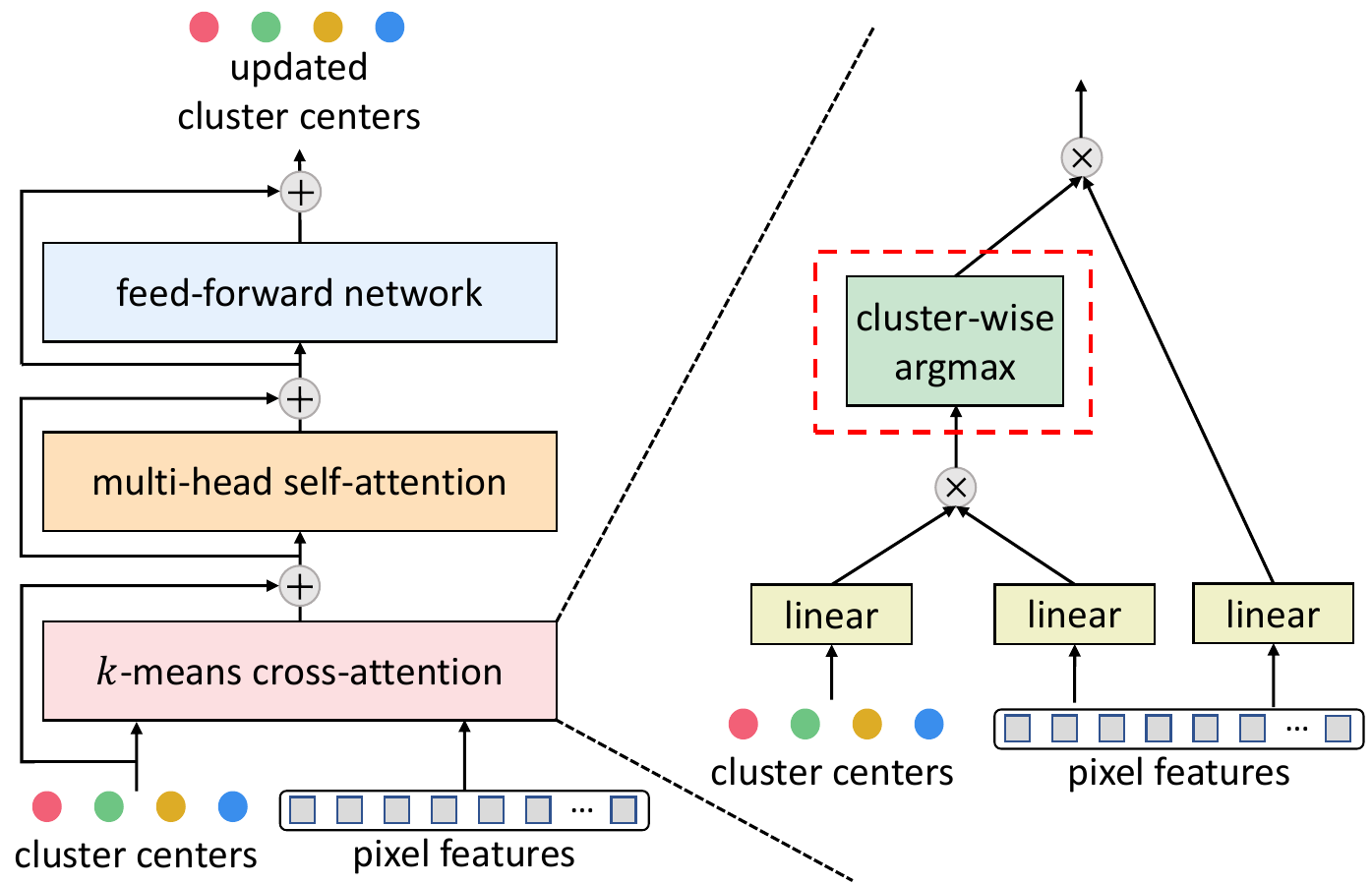}
    \caption{
    To convert a typical transformer decoder into our $k$MaX decoder, we simply replace the original cross-attention with our $k$-means cross-attention (\ie, with the only simple change \textit{cluster-wise argmax} high-lighted in red)
    }
    \label{fig:kmeans_mask_decoder}
\end{figure}

Comparing~\equref{eq:detr_transformer_update} and~\equref{eq:kmeans_attention_update}, the spatial-wise softmax is now replaced by the cluster-wise argmax.
As shown in \figref{fig:kmeans_mask_decoder}, with such a simple yet effective change, a typical transformer decoder could be converted to a $k$MaX decoder. 
Unlike the original cross-attention, the proposed $k$-means cross-attention adopts a different operation (\ie, cluster-wise argmax) to compute the attention map, and does not require the multi-head mechanism~\cite{vaswani2017attention}.
However, the cluster-wise argmax, as a hard assignment to aggregate pixel features for the cluster center update, is not a differentiable operation, posing a challenge during training. We have explored several methods (\eg, Gumbel-Softmax~\cite{jang2017categorical}), and discover that a simple deep supervision scheme turns out to be most effective. In particular, in our formulation, the affinity logits between pixel features and cluster centers directly correspond to the softmax logits of segmentation masks (\ie, $\mathbf{Q}^{c} \times (\mathbf{K}^{p})^{\mathrm{T}}$ in~\equref{eq:kmeans_attention_update} corresponds to $\mathbf{F} \times \mathbf{C}^{\mathrm{T}}$ in~\equref{eq:detr_final_output}), since the cluster centers aim to group pixels of similar affinity together to form the predicted segmentation masks. This formulation allows us to add deep supervision to every $k$MaX decoder, in order to train the parameters in  the $k$-means cross-attention module.

\begin{figure}[!t]
    \centering
    \includegraphics[width=0.69\linewidth]{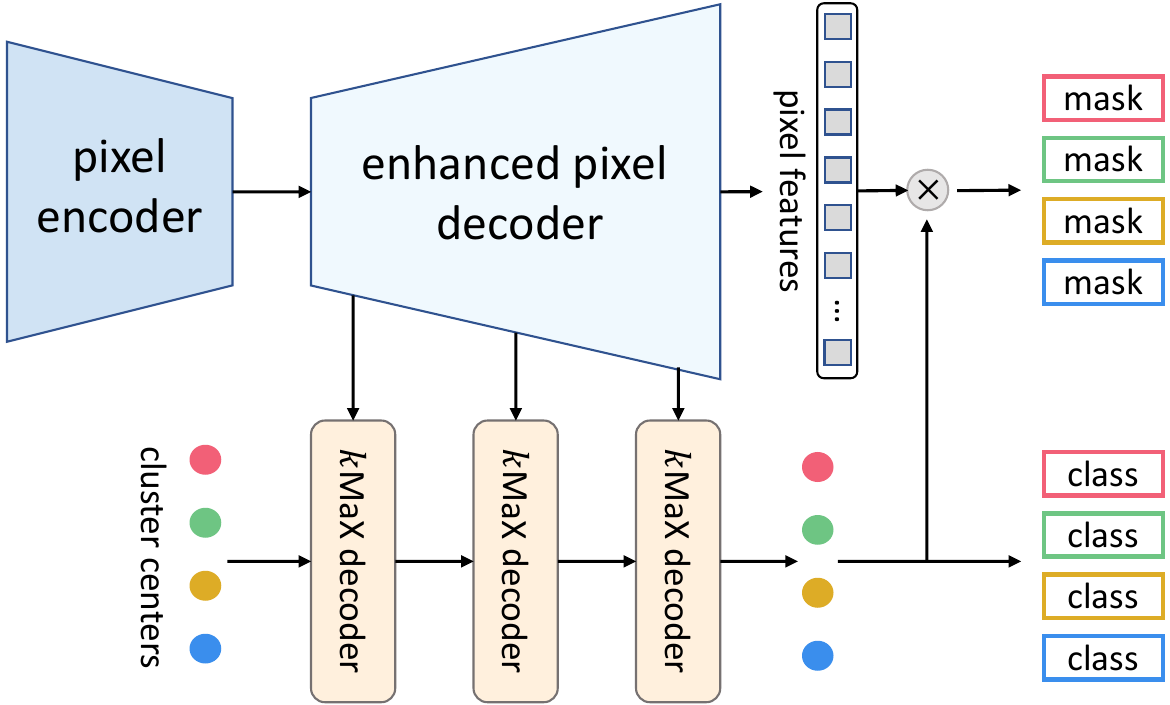}
    \caption{The meta architecture of $k$-means Mask Transformer consists of three components: pixel encoder, enhanced pixel decoder, and $k$MaX decoder. The pixel encoder is any network backbone. The enhanced pixel decoder includes transformer encoders to enhance the pixel features, and upsampling layers to generate higher resolution features. The series of $k$MaX decoders transform cluster centers into (1) mask embedding vectors, which multiply with the pixel features to generate the predicted masks, and (2) class predictions for each mask.
    }
    \label{fig:meta_arch}
\end{figure}

{\bf Meta architecture.}\quad
\figref{fig:meta_arch} shows the meta architecture of our proposed $k$MaX-DeepLab, which contains three main components: pixel encoder, enhanced pixel decoder, and $k$MaX decoder. The pixel encoder extracts the pixel features either by a CNN~\cite{he2016deep} or a transformer~\cite{liu2021swin} backbone, while the enhanced pixel decoder is responsible for recovering the feature map resolution as well as enhancing the pixel features via transformer encoders~\cite{vaswani2017attention} or axial attention~\cite{wang2020axial}. Finally, the $k$MaX decoder transforms the object queries (\ie, cluster centers) into mask embedding vectors from the $k$-means clustering perspective.

\begin{figure}[t]
    \centering
    \includegraphics[width=0.9\linewidth]{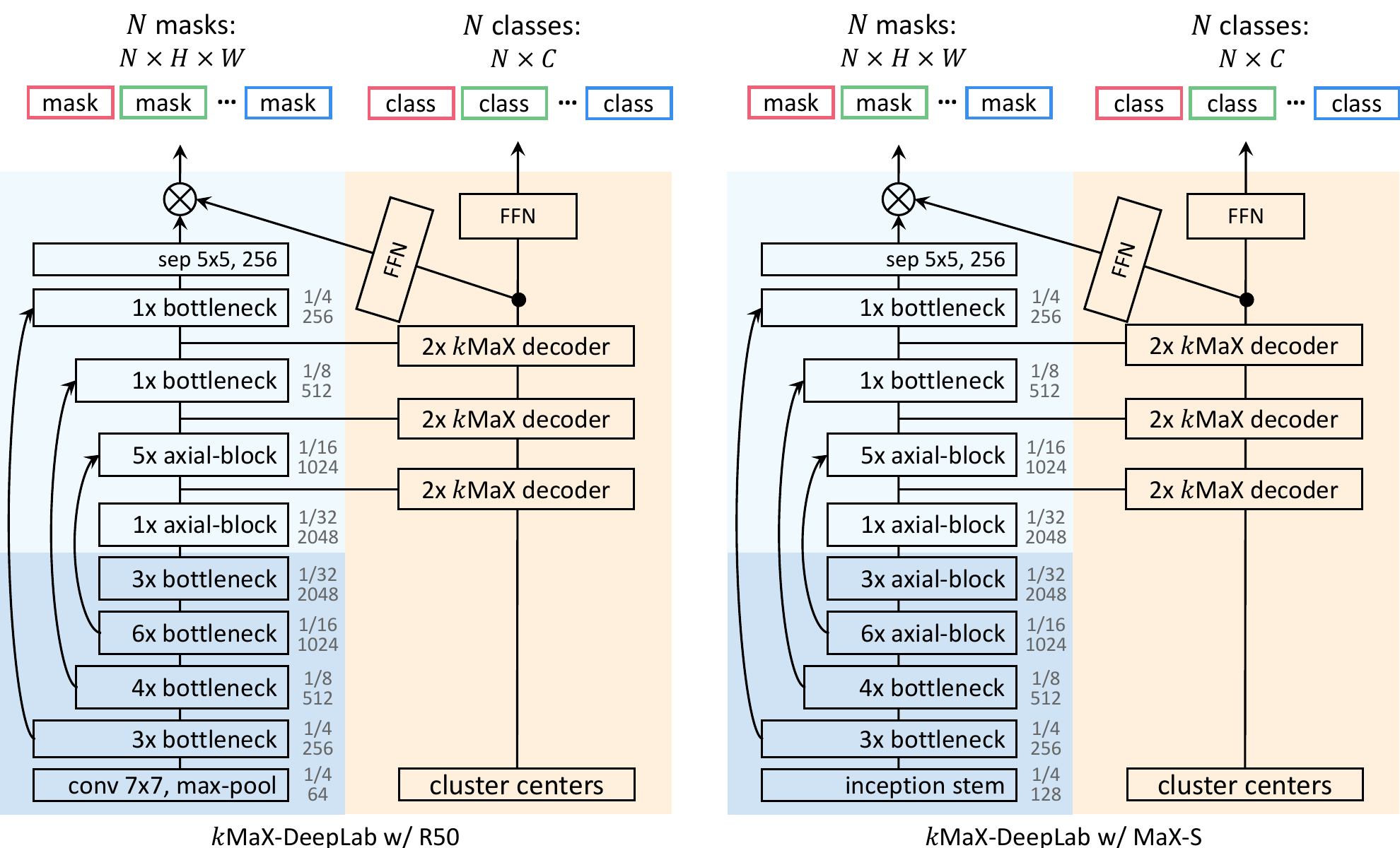}
    \caption{An illustration of $k$MaX-DeepLab with ResNet-50 and MaX-S as backbones. The hidden dimension of FFN is 256. The design of $k$MaX-DeepLab is general to different backbones by simply updating the pixel encoder (marked in dark-blue). The enhanced pixel decoder and $k$MaX decoder are colored in light-blue and yellow, respectively}
    \label{fig:detailed_kmax}
\end{figure}

{\bf Model instantiation.}\quad
We build $k$MaX based on MaX-DeepLab~\cite{wang2021max} with the official code-base~\cite{deeplab2_2021}. We divide the whole model into two paths: the pixel path and the cluster path, which are responsible for extracting pixel features and cluster centers, respectively. \figref{fig:detailed_kmax} details our $k$MaX-DeepLab instantiation with two example backbones.

\textbf{Pixel path.}\quad
The pixel path consists of a pixel encoder and an enhanced pixel decoder. The pixel encoder is an ImageNet-pretrained~\cite{russakovsky2015imagenet} backbone, such as ResNet~\cite{he2016deep}, MaX-S~\cite{wang2021max} (\ie, ResNet-50 with axial attention~\cite{wang2020axial}), and ConvNeXt~\cite{liu2022convnet}. Our enhanced pixel decoder consists of several axial attention blocks~\cite{wang2020axial} and bottleneck blocks~\cite{he2016deep}.

\textbf{Cluster path.}\quad
The cluster path contains totally six $k$MaX decoders, which are evenly distributed among features maps of different spatial resolutions. Specifically, we deploy two $k$MaX decoders each for pixel features at output stride 32, 16, and 8, respectively.

\textbf{Loss functions.}\quad
Our training loss functions mostly follow the setting of MaX-DeepLab~\cite{wang2021max}.
We adopt the same PQ-style loss, auxiliary semantic loss, mask-id cross-entropy loss, and pixel-wise instance discrimination loss~\cite{yu2022cmt}.

\label{sec:method}

\section{Experimental Results}

In this section, we first provide our implementation details. We report our main results on COCO~\cite{lin2014microsoft}, Cityscapes~\cite{Cordts2016Cityscapes}, and ADE20K~\cite{zhou2017scene}. We also provide visualizations to better understand the clustering process of the proposed $k$MaX-DeepLab.
The ablation studies are provided in the appendix.

\subsection{Implementation Details}

The meta architecture of the proposed $k$MaX-DeepLab contains three main components: the pixel encoder, enhanced pixel decoder, and $k$MaX decoder, as shown in~\figref{fig:meta_arch}. We provide the implementation details of each component below.

\textbf{Pixel encoder.}\quad
The pixel encoder extracts pixel features given an image.
To verify the generality of $k$MaX-DeepLab across different pixel encoders, we experiment with ResNet-50~\cite{he2016deep}, MaX-S~\cite{wang2021max} (\ie, ResNet-50 with axial attention~\cite{wang2020axial} in the 3rd and 4th stages), and ConvNeXt~\cite{liu2022convnet}.

\textbf{Enhanced pixel decoder.}\quad
The enhanced pixel decoder recovers the feature map resolution and  enriches pixel features via self-attention.
As shown in~\figref{fig:detailed_kmax}, we adopt one axial block with channels 2048 at output stride 32, and five axial blocks with channels 1024 at output stride 16. The axial block is a bottleneck block~\cite{he2016deep}, but the $3 \times 3$ convolution is replaced by the axial attention~\cite{wang2020axial}.
We use one bottleneck block at output stride 8 and 4, respectively.
We note that the axial blocks play the same role (\ie, feature enhancement) as the transformer encoders in other works~\cite{carion2020end,cheng2021per,yu2022cmt}, where we ensure that the total number of axial blocks is six for a fair comparison to previous works~\cite{carion2020end,cheng2021per,yu2022cmt}. 

\textbf{Cluster path.}\quad
As shown in~\figref{fig:detailed_kmax}, we deploy six $k$MaX decoders, where each two are placed for pixel features (enhanced by the pixel decoders) with output stride 32, 16, 8, respectively.
Our design uses six transformer decoders, aligning with the previous works~\cite{carion2020end,cheng2021per,yu2022cmt}, though some recent works~\cite{cheng2021masked,li2021panoptic} adopt more transformer decoders to achieve a stronger performance.

\textbf{Training and testing.}\quad
We mainly follow MaX-DeepLab~\cite{wang2021max} for training settings. The ImageNet-pretrained~\cite{russakovsky2015imagenet} backbone has a learning rate multiplier 0.1. For regularization and augmentations, we adopt drop path~\cite{huang2016deep}, random color jittering~\cite{cubuk2018autoaugment}, and panoptic copy-paste augmentation, which is an extension from instance copy-paste augmentation~\cite{fang2019instaboost,ghiasi2021simple} by augmenting both `thing' and `stuff' classes. AdamW~\cite{kingma2014adam,loshchilov2017decoupled}
optimizer is used with weight decay 0.05.
The $k$-means cross-attention adopts cluster-wise argmax, which aligns the formulation of attention map to segmentation result.
It therefore allows us to directly apply deep supervision on the attention maps.
These auxiliary losses attached to each $k$MaX decoder have the same loss weight of 1.0 as the final prediction, and Hungarian matching result based on the final prediction is used to assign supervisions for all auxiliary outputs.
During inference, we adopt the same mask-wise merging scheme used in~\cite{cheng2021per,zhang2021k,li2021panoptic,yu2022cmt} to obtain the final segmentation results. 

\textbf{COCO dataset.}\quad If not specified, we train all models with  batch size 64 on 32 TPU cores with 150k iterations (around 81 epochs). The first 5k steps serve as the warm-up stage, where the learning rate linearly increases from 0 to $5\times 10^{-4}$. The input images are resized and padded to $1281\times 1281$.
Following MaX-DeepLab~\cite{wang2021max}, the loss weights for PQ-style loss, auxiliary semantic loss, mask-id cross-entropy loss, instance discrimination loss are 3.0, 1.0, 0.3, and 1.0, respectively. The number of cluster centers (\ie, object queries) is 128, and the final feature map resolution has output stride 4 as in MaX-DeepLab~\cite{wang2021max}.

We have also experimented with doubling the number of object queries to 256 for $k$MaX-DeepLab with ConvNeXt-L, which however leads to a performance loss. Empirically, we adopt a \textbf{drop query} regularization, where we randomly drop half of the object queries (\ie, 128) during each training iteration, and all queries (\ie, 256) are used during inference. With the proposed drop query regularization, doubling the number of object queries to 256 consistently brings 0.1\% PQ improvement under the large model regime.

\textbf{Cityscapes dataset.}\quad
We train all models with batch size 32 on 32 TPU cores with 60k iterations. The first 5k steps serve as the warm-up stage, where learning rate linearly increases from 0 to $3\times 10^{-4}$. The inputs are padded to $1025\times 2049$.
The loss weights for PQ-style loss, auxiliary semantic loss, mask-id cross-entropy loss, and instance discrimination loss are 3.0, 1.0, 0.3, and 1.0, respectively.
We use 256 cluster centers, and add an additional bottleneck block in the pixel decoder to produce features with output stride 2.

\textbf{ADE20K dataset.}\quad
We adopt the same setting as the COCO dataset, except that the model is trained for 100k iterations. We experiment with both $641\times 641$ and $1281\times 1281$ input resolutions. Our inference is run \wrt. the whole input image, instead of in the sliding window manner (which may yield a better performance at the cost of more FLOPs).

\subsection{Main Results}
Our main results on the COCO~\cite{lin2014microsoft},  Cityscapes~\cite{Cordts2016Cityscapes}, and
ADE20K~\cite{zhou2017scene} \textit{val} set are summarized in~\tabref{tab:coco_val_test}, ~\tabref{tab:cityscapes_val_test}, and~\tabref{tab:ade20k_val}, respectively.

\begin{table*}[!t]
\centering
\small
\caption{COCO \textit{val} set results. Our FLOPs and FPS are evaluated with the input size $1200\times 800$ and a Tesla V100-SXM2 GPU. $\dagger$: ImageNet-22K pretraining. $\star$: Using 256 object queries with drop query regularization. $\ddagger$: Using COCO \textit{unlabeled} set
}
%\vspace{1ex}
\tablestyle{1pt}{0.9}
\begin{tabular}{l|c|ccc|ccc}
method & backbone & params & FLOPs & FPS & PQ & PQ\textsuperscript{Th} & PQ\textsuperscript{St}\\
\shline
MaskFormer~\cite{cheng2021per} & ResNet-50~\cite{he2016deep} & 45M & 181G & 17.6 & 46.5 & 51.0 & 39.8 \\
K-Net~\cite{zhang2021k} &  ResNet-50~\cite{he2016deep} & - & - & - & 47.1 & 51.7 & 40.3 \\
CMT-DeepLab~\cite{yu2022cmt}  &  ResNet-50~\cite{he2016deep} & - & - & - & 48.5 & - & - \\
Panoptic SegFormer~\cite{li2021panoptic}  &  ResNet-50~\cite{he2016deep} & 51M & 214G & 7.8 & 49.6 & 54.4 & 42.4 \\
Mask2Former~\cite{cheng2021masked}  &  ResNet-50~\cite{he2016deep} & 44M & 226G & 8.6 & 51.9 & 57.7 & 43.0 \\
\textbf{kMaX-DeepLab} & ResNet-50~\cite{he2016deep} & 57M & 168G & 22.8 & \textbf{53.0} & \textbf{58.3} & \textbf{44.9} \\
\hline
MaX-DeepLab~\cite{wang2021max} & MaX-S~\cite{wang2021max} & 62M & 324G & - & 48.4 & 53.0 & 41.5 \\
CMT-DeepLab & MaX-S$^\dagger$~\cite{wang2021max} & 95M & 396G & 8.1 & 53.0 & 57.7 & 45.9 \\
\textbf{kMaX-DeepLab} & MaX-S$^\dagger$~\cite{wang2021max} & 74M & 240G & 16.9 & \textbf{56.2} & \textbf{62.2} & \textbf{47.1} \\
\hline
MaskFormer~\cite{cheng2021per} & Swin-B (W12)$^\dagger$~\cite{liu2021swin} & 102M & 411G & 8.4 & 51.8 & 56.9 & 44.1 \\
CMT-DeepLab~\cite{yu2022cmt} & Axial-R104$^\dagger$~\cite{yu2022cmt} & 135M & 553G & 6.0 & 54.1 & 58.8 & 47.1 \\
Panoptic SegFormer~\cite{li2021panoptic} & PVTv2-B5$^\dagger$~\cite{wang2021pvtv2} & 105M & 349G & - & 55.4 & 61.2 & 46.6 \\
Mask2Former~\cite{cheng2021masked} & Swin-B (W12)$^\dagger$~\cite{liu2021swin} & 107M & 466G & - & 56.4 & 62.4 & 47.3 \\
\textbf{kMaX-DeepLab} & ConvNeXt-B$^\dagger$~\cite{liu2022convnet} & 122M & 380G & 11.6 & \textbf{57.2} & \textbf{63.4} & \textbf{47.8} \\
\hline
MaX-DeepLab~\cite{wang2021max} & MaX-L~\cite{wang2021max} & 451M & 3692G & - & 51.1 & 57.0 & 42.2 \\
MaskFormer~\cite{cheng2021per} & Swin-L (W12)$^\dagger$~\cite{liu2021swin} & 212M & 792G & 5.2 & 52.7 & 58.5 & 44.0  \\
K-Net~\cite{zhang2021k} & Swin-L (W7)$^\dagger$~\cite{liu2021swin} & - & - & - & 54.6 & 60.2 & 46.0 \\
CMT-DeepLab~\cite{yu2022cmt} & Axial-R104-RFN$^\dagger$~\cite{qiao2021detectors} & 270M & 1114G & 3.2 & 55.3 & 61.0 & 46.6 \\
Panoptic SegFormer~\cite{li2021panoptic} & Swin-L (W7)$^\dagger$~\cite{liu2021swin} & 221M & 816G & - & 55.8 & 61.7 &  46.9 \\
Mask2Former~\cite{cheng2021masked} & Swin-L (W12)$^\dagger$~\cite{liu2021swin} & 216M & 868G & 4.0 & 57.8 & 64.2 &  48.1 \\
\textbf{kMaX-DeepLab} & ConvNeXt-L$^\dagger$~\cite{liu2022convnet} & 232M & 744G & 6.7 & 57.9 & 64.0 & 48.6 \\
\textbf{kMaX-DeepLab}$^\star$ & ConvNeXt-L$^\dagger$~\cite{liu2022convnet} & 232M & 749G & 6.6 & 58.0 & 64.2 & 48.6 \\
\textbf{kMaX-DeepLab}$^\ddagger$ & ConvNeXt-L$^\dagger$~\cite{liu2022convnet} & 232M & 744G & 6.7 & \textbf{58.1} & \textbf{64.3} & \textbf{48.8} \\
\end{tabular}
\label{tab:coco_val_test}
\end{table*}

\textbf{COCO \textit{val} set.}\quad
In~\tabref{tab:coco_val_test}, we compare our $k$MaX-DeepLab with other transformer-based panoptic segmentation methods on COCO \textit{val} set. Notably, with a simple ResNet-50 backbone, $k$MaX-DeepLab already achieves 53.0\% PQ, surpassing \textit{most} prior arts with stronger backbones. Specifically, $k$MaX-DeepLab outperforms MaskFormer~\cite{cheng2021per} and K-Net~\cite{zhang2021k},
all with the ResNet-50 backbone as well, by a large margin of \textbf{6.5\%} and \textbf{5.9\%},
while maintaining a similar level of computational costs.
Our $k$MaX-DeepLab with ResNet-50 even surpasses the largest variants of MaX-DeepLab~\cite{wang2021max} by \textbf{1.9\%} PQ (while using \textbf{7.9$\times$} fewer parameters and \textbf{22.0$\times$} fewer FLOPs), and MaskFormer (while using \textbf{3.7$\times$} fewer parameters and \textbf{4.7$\times$} fewer FLOPs) by 0.3\% PQ, respectively.
With a stronger backbone MaX-S~\cite{wang2021max}, $k$MaX-DeepLab boosts the performance to 56.2\% PQ, outperforming MaX-DeepLab with the same backbone by \textbf{7.8\%} PQ.
Our $k$MaX-DeepLab with MaX-S backbone also improves over the previous state-of-art K-Net with Swin-L~\cite{liu2021swin} by \textbf{1.6\%} PQ.
To further push the envelope, we adopt the modern CNN backbone ConvNeXt~\cite{liu2022convnet} and set new state-of-the-art results of 57.2\% PQ with ConvNeXt-B and 58.0\% PQ with ConvNeXt-L, outperforming K-Net with Swin-L by a significant margin of \textbf{3.4\%} PQ.

When compared to more recent works (CMT-DeepLab~\cite{yu2022cmt}, Panoptic SegFormer~\cite{li2021panoptic}, and Mask2Former~\cite{cheng2021masked}), $k$MaX-DeepLab still shows great performances without the advanced modules, such as deformable attention~\cite{zhu2020deformable}, cascaded transformer decoder~\cite{cheng2021masked}, and uncertainty-based pointly supervision~\cite{kirillov2020pointrend}. As different backbones are utilized for each method (\eg, PVTv2~\cite{wang2021pvtv2}, Swin~\cite{liu2021swin}, and ConvNeXt~\cite{liu2022convnet}), we start with a fair comparison using the ResNet-50 backbone. Our $k$MaX-DeepLab with ResNet-50 achieves a significant better performance compared to CMT-DeepLab, Panoptic SegFormer and Mask2Former  by a large margin of \textbf{4.5\%}, \textbf{3.4\%}, and \textbf{1.1\%} PQ, respectively.
Additionally, our model runs almost \textbf{3$\times$} faster than them (since $k$MaX-DeepLab enjoys a simple design without deformable attention).
When employing stronger backbones, $k$MaX-DeepLab with ConvNeXt-B outperforms CMT-DeepLab with Axial-R104, Panoptic SegFormer with PVTv2-B5, and Mask2Former with Swin-B (window size 12) by \textbf{3.1\%}, \textbf{1.8\%}, and \textbf{0.8\%} PQ, respectively, while all models have a similar level of cost (parameters and FLOPs).
When scaling up to the largest backbone for each method, $k$MaX-DeepLab outperforms CMT-DeepLab, and Panoptic SegFormer significantly by \textbf{2.7\%} and \textbf{2.2\%} PQ.
Although we already perform better than Mask2Former with Swin-L (window size 12), we notice that $k$MaX-DeepLab benefits much less than Mask2Former when scaling up from base model to large model (+0.7\% for $k$MaX-DeepLab but +1.4\% for Mask2Former), indicating $k$MaX-DeepLab's strong representation ability and that it may overfit on COCO \textit{train} set with the largest backbone.
Therefore, we additionally perform a simple experiment to alleviate the over-fitting issue by generating pseudo labels~\cite{chen2020naive} on COCO \textit{unlabeled} set.
Adding pseudo labels to the training data slightly improves $k$MaX-DeepLab, yielding a PQ score of \textbf{58.1\%} (the drop query regularization is not used here and the number of object query remains 128).

\textbf{Cityscapes \textit{val} set.}\quad
In~\tabref{tab:cityscapes_val_test}, we compare our $k$MaX-DeepLab with other state-of-art methods on Cityscapes \textit{val} set. Our reported PQ, AP, and mIoU results use the same panoptic model to provide a comprehensive comparison.
Notably, $k$MaX-DeepLab with ResNet-50 backbone already surpasses most baselines, while being more efficient.
For example, $k$MaX-DeepLab with ResNet-50 achieves \textbf{1.3\%} PQ higher performance compared to Panoptic-DeepLab~\cite{cheng2019panoptic} (Xception-71~\cite{chollet2016xception} backbone) with \textbf{20\%} computational cost (FLOPs) reduced.
Moreover, it achieves a similar performance to Axial-DeepLab-XL~\cite{wang2020axial}, while using \textbf{3.1$\times$} fewer parameters and \textbf{5.6$\times$} fewer FLOPs.
$k$MaX-DeepLab achieves even higher performances with stronger backbones.
Specifically, with MaX-S backbone, it performs on par with previous state-of-the-art Panoptic-DeepLab with SWideRNet~\cite{swidernet_2020} backbone, while using \textbf{7.2$\times$} fewer parameters and \textbf{17.2$\times$} fewer FLOPs.
Additionally, even only trained with panoptic annotations, our $k$MaX-DeepLab also shows superior performance in instance segmentation (AP) and semantic segmentation (mIoU).
Finally, we provide a comparison with the recent work Mask2Former~\cite{cheng2021masked}, where the advantage of our $k$MaX-DeepLab becomes even more significant.
Using the ResNet-50 backbone for a fair comparison, $k$MaX-DeepLab achieves \textbf{2.2\%} PQ, \textbf{1.2\%} AP, and \textbf{2.2\%} mIoU higher performance than Mask2Former.
For other backbone variants with a similar  size, $k$MaX-DeepLab with ConvNeXt-B is \textbf{1.9\%} PQ higher than Mask2Former with Swin-B (window size 12). Notably, $k$MaX-DeepLab with ConvNeXt-B already obtains a PQ score that is \textbf{1.4\%} higher than Mask2Former with their best backbone. With ConvNeXt-L as backbone, $k$MaX-DeepLab sets a new state-of-the-art record of 68.4\% PQ without any test-time augmentation or COCO~\cite{lin2014microsoft}/Mapillary Vistas~\cite{neuhold2017mapillary} pretraining.

\begin{table*}[!t]
\centering
\small
\caption{Cityscapes \textit{val} set results. We only consider methods without extra data~\cite{lin2014microsoft,neuhold2017mapillary} and test-time augmentation for a fair comparison. We evaluate FLOPs and FPS with the input size $1025\times 2049$ and a Tesla V100-SXM2 GPU. Our instance (AP) and semantic (mIoU) results are based on the same panoptic model (\ie, no task-specific fine-tuning). $\dagger$: ImageNet-22K pretraining
}
%\vspace{1ex}
\tablestyle{0.75pt}{0.75}
\begin{tabular}{l|c|ccc|ccc}
method & backbone & params & FLOPs & FPS & PQ & AP & mIoU\\
\shline
Panoptic-DeepLab~\cite{cheng2019panoptic} & Xception-71~\cite{chollet2016xception} & 47M & 548G & 5.7 & 63.0 & 35.3 & 80.5 \\
Axial-DeepLab~\cite{wang2020axial} & Axial-ResNet-L~\cite{wang2020axial} & 45M & 687G & - & 63.9 & 35.8 & 81.0 \\
Axial-DeepLab~\cite{wang2020axial} & Axial-ResNet-XL~\cite{wang2020axial} & 173M & 2447G & - & 64.4 & 36.7 & 80.6 \\
CMT-DeepLab~\cite{yu2022cmt} & MaX-S~\cite{wang2021max} & - & - & - & 64.6 & - & 81.4 \\
Panoptic-DeepLab~\cite{cheng2019panoptic} & SWideRNet-(1,1,4.5)~\cite{swidernet_2020} & 536M & 10365G & 1.0 & 66.4 & 40.1 & 82.2 \\
\hline
Mask2Former~\cite{cheng2021masked} & ResNet-50~\cite{he2016deep} & - & - & - & 62.1 & 37.3 & 77.5 \\
Mask2Former~\cite{cheng2021masked} & Swin-B (W12)$^\dagger$~\cite{liu2021swin} & - & - & -  & 66.1 & 42.8 & 82.7 \\
Mask2Former~\cite{cheng2021masked} & Swin-L (W12)$^\dagger$~\cite{liu2021swin} & - & - & -  & 66.6 & 43.6 & 82.9 \\
\hline
SETR~\cite{zheng2021rethinking} & ViT-L$^\dagger$~\cite{dosovitskiy2020image} & - & - & - & - & - & 79.3 \\
SegFormer~\cite{xie2021segformer} & MiT-B5~\cite{xie2021segformer} & 85M & 1460G & 2.5 & - & - & 82.4 \\
\hline
Mask R-CNN~\cite{he2017mask} & ResNet-50~\cite{he2016deep} & - & - & - & - & 31.5 & - \\
PANet~\cite{liu2018path} & ResNet-50~\cite{he2016deep} & - & - & - & - & 36.5 & - \\
\hline \hline
\textbf{kMaX-DeepLab} & ResNet-50~\cite{he2016deep} & 56M & 434G & 9.0 & 64.3 & 38.5 & 79.7 \\
\textbf{kMaX-DeepLab} & MaX-S$^\dagger$~\cite{wang2021max} & 74M & 602G & 6.5 & 66.4 & 41.6 & 82.1 \\
\textbf{kMaX-DeepLab}  & ConvNeXt-B$^\dagger$~\cite{liu2022convnet} & 121M & 858G & 5.2 & 68.0 & 43.0 & 83.1 \\
\textbf{kMaX-DeepLab} & ConvNeXt-L$^\dagger$~\cite{liu2022convnet} & 232M & 1673G & 3.1 & \textbf{68.4} & \textbf{44.0} & \textbf{83.5} \\
\end{tabular}
\label{tab:cityscapes_val_test}
\end{table*}

\textbf{ADE20K \textit{val} set.}\quad
In~\tabref{tab:ade20k_val}, we summarize $k$MaX-DeepLab's performance on ADE20K against other state-of-the-art methods. We report PQ and mIoU results with the same panoptic model, where $k$MaX-DeepLab consistently shows better performance.
Specifically, with ResNet-50 and input size $641\times641$, $k$MaX-DeepLab attains 41.5\% PQ. Increasing the input size to $1281\times1281$  further improves the performance to 42.3\% PQ, significantly outperforming the prior state-of-the-art Mask2Former with ResNet-50 backbone by 2.6\% PQ.
Finally, $k$MaX-DeepLab equipped with the modern ConvNeXt-L backone achieves a new state-of-the-art performance of 50.9\% PQ, signicantly surpassing MaskFormer, Panoptic-DeepLab with SWideRNet, and Mask2Former by 16.2\%, 13.0\%, and 2.8\% PQ, respectively.

\begin{table*}[!t]
\centering
\small
\caption{ADE20K \textit{val} set results. Our FLOPs and FPS are evaluated with the input size ($641\times 641$ or $1281\times 1281$) and a Tesla V100-SXM2 GPU. $\dagger$: ImageNet-22K pretraining. The input size for $k$MaX-DeepLab is shown in the parentheses 
}
%\vspace{1ex}
\tablestyle{1pt}{0.9}
\begin{tabular}{l|c|ccc|cc}
method & backbone & params & FLOPs & FPS & PQ  & mIoU \\
\shline
MaskFormer~\cite{cheng2021per} & ResNet-50~\cite{he2016deep} & - & - & - & 34.7 & - \\
Panoptic-DeepLab~\cite{cheng2019panoptic} & SWideRNet-(1,1.5,3)~\cite{swidernet_2020} & - & - & - & 37.4 & 50.4 \\
Panoptic-DeepLab~\cite{cheng2019panoptic} & SWideRNet-(1,1,4)~\cite{swidernet_2020} & - & - & - & 37.9 & 50.0 \\
Mask2Former~\cite{cheng2021masked} & ResNet-50~\cite{he2016deep} & - & - & - & 39.7 & 46.1 \\
Mask2Former~\cite{cheng2021masked} & Swin-L (W12)$^\dagger$~\cite{liu2021swin} & - & - & - & 48.1 & 54.5 \\
\hline \hline
$k$MaX-DeepLab (641) & ResNet-50~\cite{he2016deep} & 57M & 75G & 38.7 & 41.5 & 45.0 \\
$k$MaX-DeepLab (1281) & ResNet-50~\cite{he2016deep} & 57M & 295G & 14.4 & 42.3 & 45.3 \\
$k$MaX-DeepLab (641) & ConvNeXt-L$^\dagger$~\cite{liu2022convnet} & 232M & 333G & 14.0 & 48.7 & 54.8 \\
$k$MaX-DeepLab (1281) & ConvNeXt-L$^\dagger$~\cite{liu2022convnet} & 232M & 1302G & 4.0 & 50.9 & 55.2 \\
\end{tabular}
\label{tab:ade20k_val}
\end{table*}

\begin{figure}[t]
    \centering
    \includegraphics[width=1.0\linewidth]{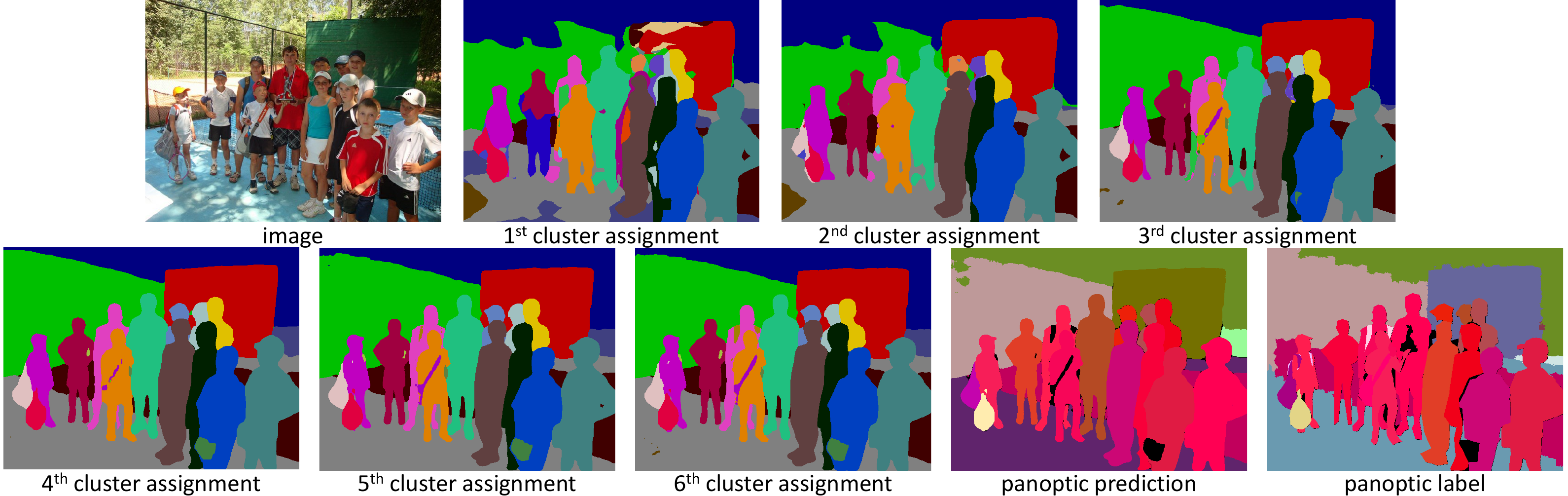}
    \caption{Visualization of $k$MaX-DeepLab (ResNet-50) pixel-cluster assignments at each $k$MaX decoder stage, along with the final panoptic prediction. In the cluster assignment visualization, pixels with same color are assigned to the same cluster and their features will be aggregated for updating corresponding cluster centers}
    \label{fig:vis_kmax}
\end{figure}

\begin{figure}[t]
    \centering
    \includegraphics[width=1.0\linewidth]{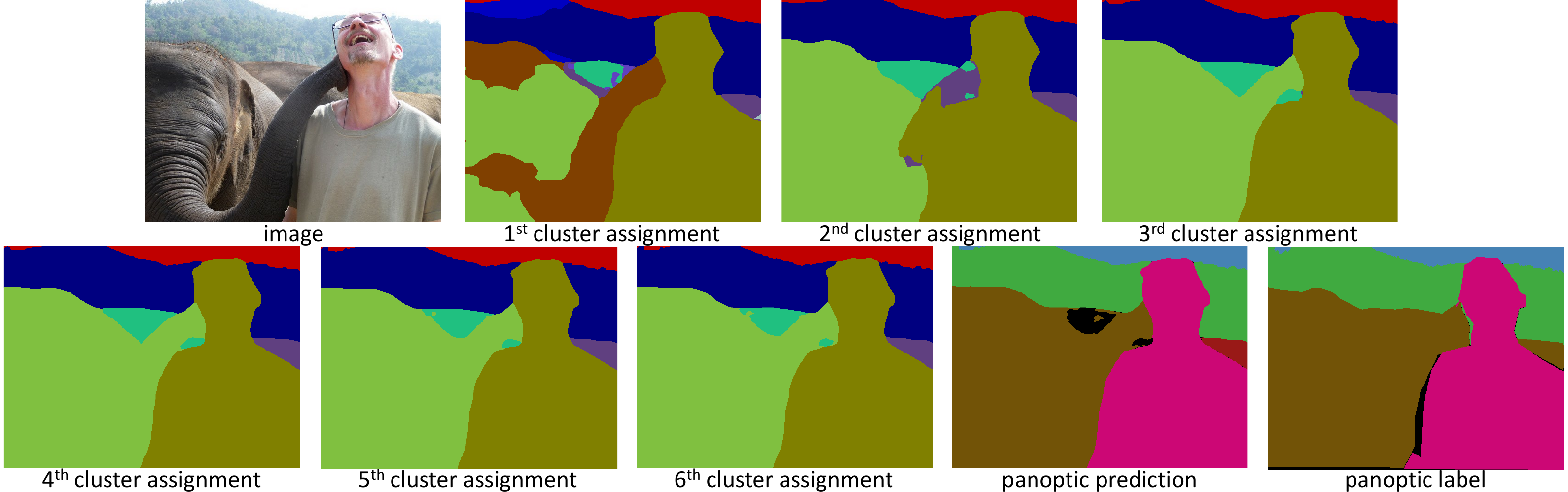}
    \caption{
    Visualization of $k$MaX-DeepLab (ResNet-50) pixel-cluster assignments at each $k$MaX decoder stage, along with the final panoptic prediction. $k$MaX-DeepLab shows a behavior of recognizing objects starting from their parts to their the whole shape in the clustering process. For example, the elephant's top head, body, and nose are separately clustered at the beginning, and they are gradually merged in the following stages}
    \label{fig:supp_vis_kmax_4}
\end{figure}

\textbf{Visualizations.}\quad 
In~\figref{fig:vis_kmax}, we provide a visualization of pixel-cluster assignments at each $k$MaX decoder and final prediction, to better understand the working mechanism behind $k$MaX-DeepLab. Another benefit of $k$MaX-DeepLab is that with the cluster-wise argmax, visualizations can be directly drawn as segmentation masks, as the pixel-cluster assignments are exclusive to each other with cluster-wise argmax. Noticeably, the major clustering update happens in the first three stages, which already updates cluster centers well and generates reasonable clustering results, while the following stages mainly focus on refining details. This coincides with our observation that 3 $k$MaX decoders are sufficient to produce good results. Besides, we observe that 1st clustering assignment tends to produce over-segmentation effects, where many clusters are activated and then combined or pruned in the later stages. Moreover, though there exist many fragments in the first round of clustering, it already surprisingly distinguishes different semantics, especially some persons are already well clustered, which indicates that the initial clustering is not only based on texture or location, but also depends on the underlying semantics. Another visualization is shown in~\figref{fig:supp_vis_kmax_4}, where we observe that $k$MaX-DeepLab behaves in a part-to-whole manner to capture an instance. More experimental results (\eg, ablation studies, test set results) and visualizations are available in the appendix.

\label{sec:results}
\section{Conclusion}
\label{sec:conclusion}
%\vspace{-1ex}
In this work, we have presented a novel end-to-end framework, called $k$-means Mask Transformer ($k$MaX-DeepLab), for segmentation tasks.
$k$MaX-DeepLab rethinks the relationship between pixel features and object queries from the clustering perspective.
Consequently, it simplifies the mask-transformer model by replacing the multi-head cross attention with the proposed single-head $k$-means clustering.
We have tailored the transformer-based model for segmentation tasks by establishing the link between the traditional $k$-means clustering algorithm and cross-attention.
We hope our work will inspire the community to develop more vision-specific transformer models.

\paragraph{Acknowledgments.}
We thank Jun Xie for the valuable feedback on the draft. This work was supported in part by ONR N00014-21-1-2812.

\clearpage
% ---- Bibliography ----
%
% BibTeX users should specify bibliography style 'splncs04'.
% References will then be sorted and formatted in the correct style.
%
\bibliographystyle{splncs04}
\bibliography{egbib}

\clearpage

\appendix
In the appendix, we provide ablation studies, along with both COCO~\cite{lin2014microsoft} and Cityscapes~\cite{Cordts2016Cityscapes} \textit{test} set results. We also include more visualizations and some failure cases.

\section{More Experimental Results}

\subsection{Ablation Studies}
We conduct ablation studies on COCO \textit{val} set. To ensure the conclusion is general to different backbones, we experiment with both ResNet-50~\cite{he2016deep} and MaX-S~\cite{wang2021max} (\ie, ResNet-50 with axial-attention blocks~\cite{wang2020axial} in the 3rd and 4th stages). Models are trained with 100k iterations for experiment efficiency.

\textbf{Different ways for pixel-cluster interaction.}\quad
The proposed $k$-means cross-attention adopts a different operation (\ie, cluster-wise argmax) from the original cross-attention (\ie, spatial-wise softmax)~\cite{vaswani2017attention}. The modification, even though simple, significantly improves the performance at a negligible cost of extra parameters and FLOPs (incurred by the extra prediction heads for deep supervision). 
In \tabref{tab:pixel_cluster_interaction_ablation}, we provide a comparison with different cross-attention modules serving for the pixel-cluster interaction.
In this ablation study, we keep everything the same (\eg, the network architecture and training recipes) except the `cross-attention modules'.
As shown in the table, $k$-means cross-attention significantly surpasses the original cross-attention by 5.2\% PQ with ResNet-50 as backbone. Even when employing a stronger backbone MaX-S, we still observe a significant gain of 4.1\% PQ. In both cases, the proposed $k$-means cross-attention maintains a similar level of parameters and FLOPs.

We have also experimented with another improved cross-attention: dual-path cross-attention, proposed in~\cite{wang2021max}.
The dual-path cross-attention simultaneously updates pixel features and cluster centers, and only shows a marginal improvement (\eg, 0.5\% with ResNet-50) over the original cross-attention at the cost of more parameters and FLOPs.
Additionally, we attempted to combine both dual-path cross-attention and the proposed $k$-means cross-attention (called dual-path $k$-means cross-attention in the table), but did not observe any further significant improvement. Therefore, we did not use it in our final model. 

Additionally, we try to add the deep supervision to the cross-attention variant as well, which degrades 1.2\% PQ for ResNet-50 backbone and improves 0.1\% PQ for MaX-S backbone, indicating that deep supervision, though needed to train the $k$MaX decoder, is not the reason of the performance improvement.

\textbf{Number of $k$MaX decoders.}\quad
In~\tabref{tab:num_of_dec_ablation}, we study the effect of deploying a different number of $k$MaX decoders at feature maps with output stride 32, 16, and 8.
For simplicity, we only experiment with using the same number of decoders for each resolution.
We note that a more complex combination is possible, but it is not the main focus of this paper.
As shown in the table, using one $k$MaX decoder at each resolution (denoted as (1, 1, 1) in the table), our $k$MaX-DeepLab already achieves a good performance of 52.5\% PQ and 55.8\% PQ with ResNet-50 and MaX-S as backbones, respectively.
Adding one more $k$MaX decoder per resolution (denoted as (2, 2, 2) in the table) further improves the performance to 52.7\% PQ and 56.1\% PQ with ResNet-50 and MaX-S as backbone, respectively.
The performance starts to saturate when using more $k$MaX decoders.
In the end, we employ totally six $k$MaX decoders, evenly distributed at output stride 32, 16, and 8 (see~\figref{fig:detailed_kmax} for a reference).

\begin{table}[!t]
\centering
\small
\caption{Ablation on different ways for pixel-cluster interaction. The final setting used in $k$MaX-DeepLab is labeled with \textcolor{gray}{gray} color
}
\tablestyle{3pt}{1.0}
\begin{tabular}{l|ccc|ccc}
 & \multicolumn{3}{c}{ResNet-50} & \multicolumn{3}{c}{MaX-S} \\
pixel-cluster interaction module & params & FLOPs & PQ & params & FLOPs & PQ \\
\shline
cross-attention~\cite{vaswani2017attention} & 56M & 165G & 47.5 & 73M & 237G & 52.0 \\
dual-path cross-attention~\cite{wang2021max} & 58M & 175G & 48.0 & 75M & 247G & 52.3 \\
\hline
\baseline{$k$-means cross-attention} & \baseline{57M} & \baseline{168G} & \baseline{52.7} & \baseline{74M} & \baseline{240G} & \baseline{56.1} \\
dual-path $k$-means cross-attention & 59M & 176G & 53.0 & 76M & 248G & 56.2 \\
\end{tabular}
\label{tab:pixel_cluster_interaction_ablation}
\end{table}

\begin{table}[t]
\centering
\small
\caption{Ablation on the number of $k$MaX decoders. The three numbers (x, y, z) of each entry in column one correspond to the number of $k$MaX decoders deployed at output stride 32, 16, and 8, respectively. For simplicity, we only experiment with using the same number of decoders for each resolution. The final setting used in $k$MaX-DeepLab is labeled with \textcolor{gray}{gray} color}
\tablestyle{4pt}{1.0}
\begin{tabular}{c|ccc|ccc}
 number of & \multicolumn{3}{c}{ResNet-50} & \multicolumn{3}{c}{MaX-S} \\
 $k$MaX decoders & params & FLOPs & PQ & params & FLOPs & PQ \\
\shline
(1, 1, 1) & 52M & 159G & 52.5 & 68M & 231G & 55.8 \\
\baseline{(2, 2, 2)} & \baseline{57M} & \baseline{168G} & \baseline{52.7} & \baseline{74M} & \baseline{240G} & \baseline{56.1} \\
(3, 3, 3) & 63M & 176G & 52.8 & 80M & 248G & 56.0 \\
\end{tabular}
\label{tab:num_of_dec_ablation}
\end{table}

\textbf{Training convergence.}\quad
As a comparison of training convergence, we train $k$MaX-DeepLab for 25k, 50k, 100k, 125k, 150k iterations, which gives 48.8\%, 51.3\%, 52.7\%, 53.0\%, and 53.0\% for ResNet-50 backbone, and 52.4\%, 54.6\%, 56.1\%, 56.1\%, 56.2\% for MaX-S backbone, respectively.
Notably, $k$MaX-DeepLab not only shows a consistent and significant improvement over its baseline MaX-DeepLab~\cite{wang2021max}, but also shows a trend to converge at 150k, while the MaX-DeepLab requires much more training iterations to converge (\eg, MaX-DeepLab with MaX-S gets 0.8\% and 1.1\% improvement when trained for 200k, 400k, respectively).

\textbf{COCO test set.}\quad
We provide comparison to prior arts on COCO \textit{test} set in \tabref{tab:coco_test}. The performance of $k$MaX-DeepLab on \textit{val} set successfully transfers to \textit{test} set. We analyze the results below \wrt different backbones.

\begin{enumerate}
    \item With ResNet-50~\cite{he2016deep}, $k$MaX-DeepLab outperforms MaX-DeepLab~\cite{wang2021max} with MaX-L by \textbf{2.1\%} PQ, while requring \textbf{7.9$\times$} fewer parameters and \textbf{22.0$\times$} fewer computation.
    \item Using MaX-S~\cite{wang2021max} backbone, $k$MaX-DeepLab outperforms  MaskFormer~\cite{cheng2021per} with Swin-L (window size 12) by \textbf{3.1\%} PQ, while requiring \textbf{2.9$\times$} fewer parameters, \textbf{3.3$\times$} fewer FLOPs, and runs \textbf{3.2$\times$} faster (FPS).
    Additionally, $k$MaX-DeepLab surpasses previous state-of-the-art method K-Net~\cite{zhang2021k} by \textbf{1.2\%} PQ.
    \item Using ConvNeXt-L~\cite{liu2022convnet} backbone, $k$MaX-DeepLab sets a new state-of-the-art result with 58.5\% PQ, significantly outperforms the best variant of MaskFormer, K-Net, and some recent works CMT-DeepLab, Panoptic SegFormer and Mask2Former by \textbf{5.2\%}, \textbf{3.3\%}, \textbf{2.8\%} PQ, \textbf{2.3\%}, and \textbf{0.2\%}, respectively. 
\end{enumerate}

\textbf{Cityscapes test set.}\quad
The Cityscapes test set results are summarized in \tabref{tab:cityscapes_test}, where our $k$MaX-DeepLab does not use any external datasets~\cite{neuhold2017mapillary,lin2014microsoft} or test-time augmentation.
We observe that $k$MaX-DeepLab, with single-scale testing, shows a significant improvement of \textbf{1.4\%} PQ compared to the previous state-of-art Panoptic-DeepLab~\cite{cheng2019panoptic} with SWideRNet-(1, 1, 4.5)~\cite{swidernet_2020} as backbone, which adopts multi-scale testing, resulting in over \textbf{60$\times$} more computational costs compared to $k$MaX-DeepLab. Finally, as shown in the table, even compared with other task-specific models, our $k$MaX-DeepLab also outperforms them in terms of instance segmentation (\textbf{1.7\%} and \textbf{7.9\%} AP over Panoptic-DeepLab and PANet~\cite{liu2018path}, respectively) and semantic segmentation (\textbf{2.8\%} and \textbf{1.0\%} mIoU better than Panoptic-DeepLab and SegFormer~\cite{xie2021segformer}, respectively).
Our reported PQ, AP, and mIoU are obtained by a single panoptic model without any task-specific fine-tuning.
This demonstrates that $k$MaX-DeepLab is a general method for different segmentation tasks.

\begin{table*}[!t]
\centering
\small
\caption{COCO \textit{test} set results. Our FLOPs and FPS are evaluated with the input size $1200\times 800$ and a Tesla V100-SXM2 GPU. $\dagger$: ImageNet-22K pretraining. $\star$: Using 256 object queries with drop query regularization. $\ddagger$: Using COCO \textit{unlabeled} set
}
%\vspace{1ex}
\tablestyle{1pt}{0.9}
\begin{tabular}{l|c|ccc|ccc}
method & backbone & params & FLOPs & FPS & PQ  & PQ\textsuperscript{Th} & PQ\textsuperscript{St}\\
\shline
MaX-DeepLab~\cite{wang2021max} & MaX-S~\cite{wang2021max} & 62M & 324G & - & 49.0 & 54.0 & 41.6 \\
MaX-DeepLab~\cite{wang2021max} & MaX-L~\cite{wang2021max} & 451M & 3692G & - & 51.3  & 57.2 & 42.4 \\
MaskFormer~\cite{cheng2021per} & Swin-L (W12)$^\dagger$~\cite{liu2021swin} & 212M & 792G & 5.2 & 53.3 & 59.1 & 44.5  \\
K-Net~\cite{zhang2021k} & Swin-L (W7)$^\dagger$~\cite{liu2021swin} & - & - & - & 55.2  & 61.2 & 46.2 \\
CMT-DeepLab~\cite{yu2022cmt} & Axial-R104-RFN$^\dagger$~\cite{qiao2021detectors} & 270M & 1114G & 3.2 & 55.7  & 61.6 & 46.8 \\
Panoptic SegFormer~\cite{li2021panoptic} & Swin-L (W7)$^\dagger$~\cite{liu2021swin} & 221M & 816G & - & 56.2 & 62.3 & 47.0 \\
Mask2Former~\cite{cheng2021masked} & Swin-L (W12)$^\dagger$~\cite{liu2021swin} & 216M & 868G & 4.0 & 58.3 & \textbf{65.1} & 48.1 \\
\hline \hline
$k$MaX-DeepLab & ResNet-50~\cite{he2016deep} & 57M & 168G & 22.8 & 53.4 & 59.3 & 44.5 \\
$k$MaX-DeepLab & MaX-S$^\dagger$~\cite{wang2021max} & 74M & 240G & 16.9 & 56.4 & 62.7 & 46.9 \\
$k$MaX-DeepLab & ConvNeXt-B$^\dagger$~\cite{liu2022convnet} & 122M & 380G & 11.6 & 57.8 & 64.3 & 48.1 \\
$k$MaX-DeepLab & ConvNeXt-L$^\dagger$~\cite{liu2022convnet} & 232M & 744G & 6.7 & 58.0 & 64.5 & 48.2 \\
$k$MaX-DeepLab$^\star$ &
ConvNeXt-L$^\dagger$~\cite{liu2022convnet} & 232M & 749G & 6.6 & 58.2 & 64.7 & 48.5 \\
$k$MaX-DeepLab$^\ddagger$ &
ConvNeXt-L$^\dagger$~\cite{liu2022convnet} & 232M & 744G & 6.7 & \textbf{58.5} & 64.8 & \textbf{49.0} \\
\end{tabular}
\label{tab:coco_test}
\end{table*}

\begin{table*}[!t]
\centering
\small
\caption{Cityscapes \textit{test} set results. $\dagger$: ImageNet-22K pretraining. TTA: test-time augmentation (which usually incurs at least 10$\times$ more computational cost). Our reported PQ, AP, and mIoU are obtained by a single panoptic model (\ie, no task-specific fine-tuning). We mainly consider results without external dataset (\eg, Mapillary Vistas, COCO) for a fair comparison}

%\vspace{1ex}
\tablestyle{1pt}{0.9}
\begin{tabular}{l|c|c|ccc}
method & backbone & TTA & PQ & AP & mIoU\\
\shline
Panoptic-DeepLab~\cite{cheng2019panoptic} & Xception-71~\cite{chollet2016xception} & \checkmark & 62.3 & 34.6 & 79.4 \\
Axial-DeepLab~\cite{wang2020axial} & Axial-ResNet-XL~\cite{wang2020axial} & \checkmark& 62.8 & 34.0 & 79.9 \\
Panoptic-DeepLab~\cite{cheng2019panoptic} & SWideRNet-(1,1,4.5)~\cite{swidernet_2020} & \checkmark & 64.8 & 38.0 & 80.4 \\
\hline
SETR~\cite{zheng2021rethinking} & ViT-L$^\dagger$~\cite{dosovitskiy2020image} & \checkmark & - & - & 81.1 \\
SegFormer~\cite{xie2021segformer} & MiT-B5~\cite{xie2021segformer} & \checkmark & - & - & 82.2 \\
\hline
Mask R-CNN~\cite{he2017mask} & ResNet-50~\cite{he2016deep} &  & - & 26.2 & - \\
PANet~\cite{liu2018path} & ResNet-50~\cite{he2016deep} & & - & 31.8 & - \\
\hline \hline
$k$MaX-DeepLab & ConvNeXt-L$^\dagger$~\cite{liu2022convnet} &  & 66.2 & 39.7 & 83.2 \\
\end{tabular}
\label{tab:cityscapes_test}
\end{table*}

\section{Visualization}
\begin{figure}[b]
    \centering
    \includegraphics[width=1.0\linewidth]{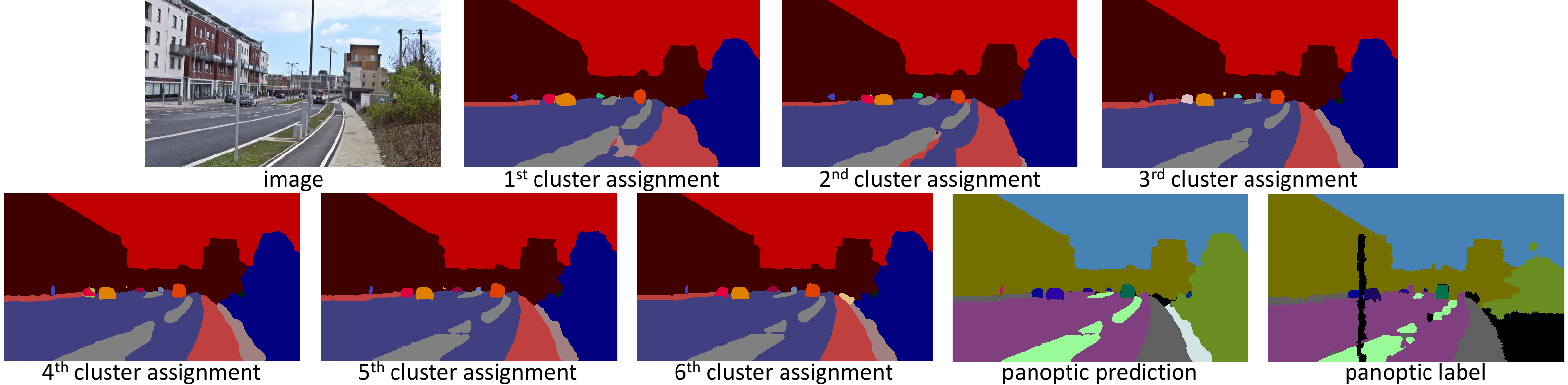}
    \caption{
    $k$MaX-DeepLab is capable of capturing extremely small objects, which may be even missing in the ground truth annotation (\eg, the person on the street, in the left side of the image. Best viewed zoom in)}
    \label{fig:supp_vis_kmax_3}
\end{figure}

\begin{figure}[t]
    \centering
    \includegraphics[width=1.0\linewidth]{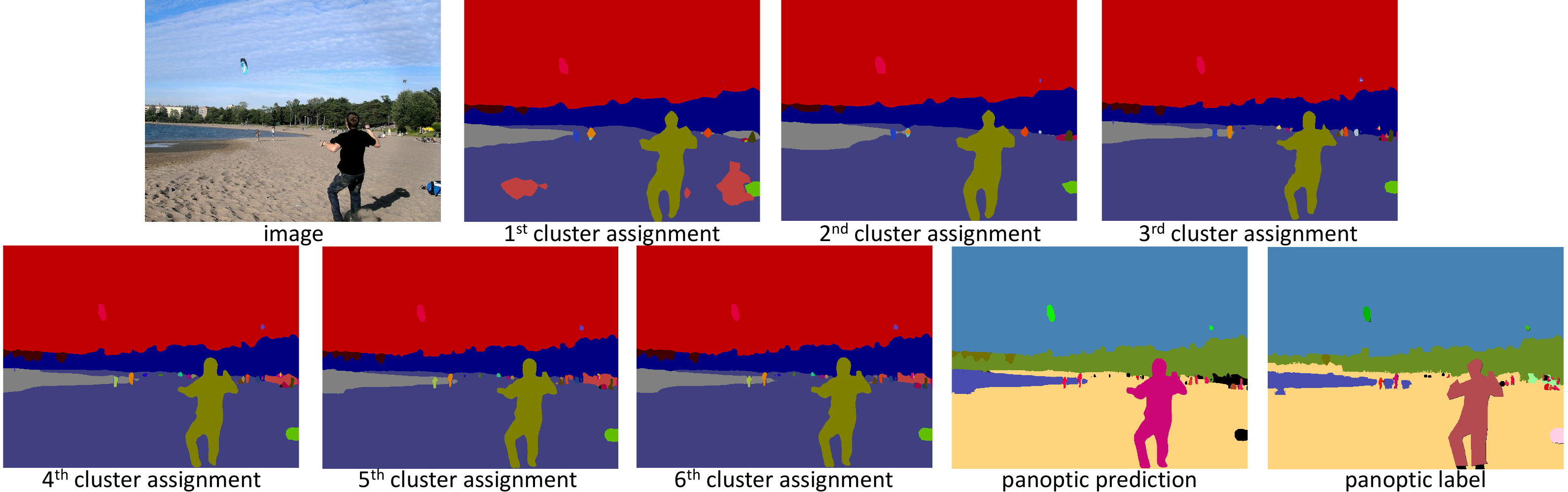}
    \caption{
    $k$MaX-DeepLab is capable of handling images with many small objects in a complex scene
    }
    \label{fig:supp_vis_kmax_5}
\end{figure}

\begin{figure}[t]
    \centering
    \includegraphics[width=1.0\linewidth]{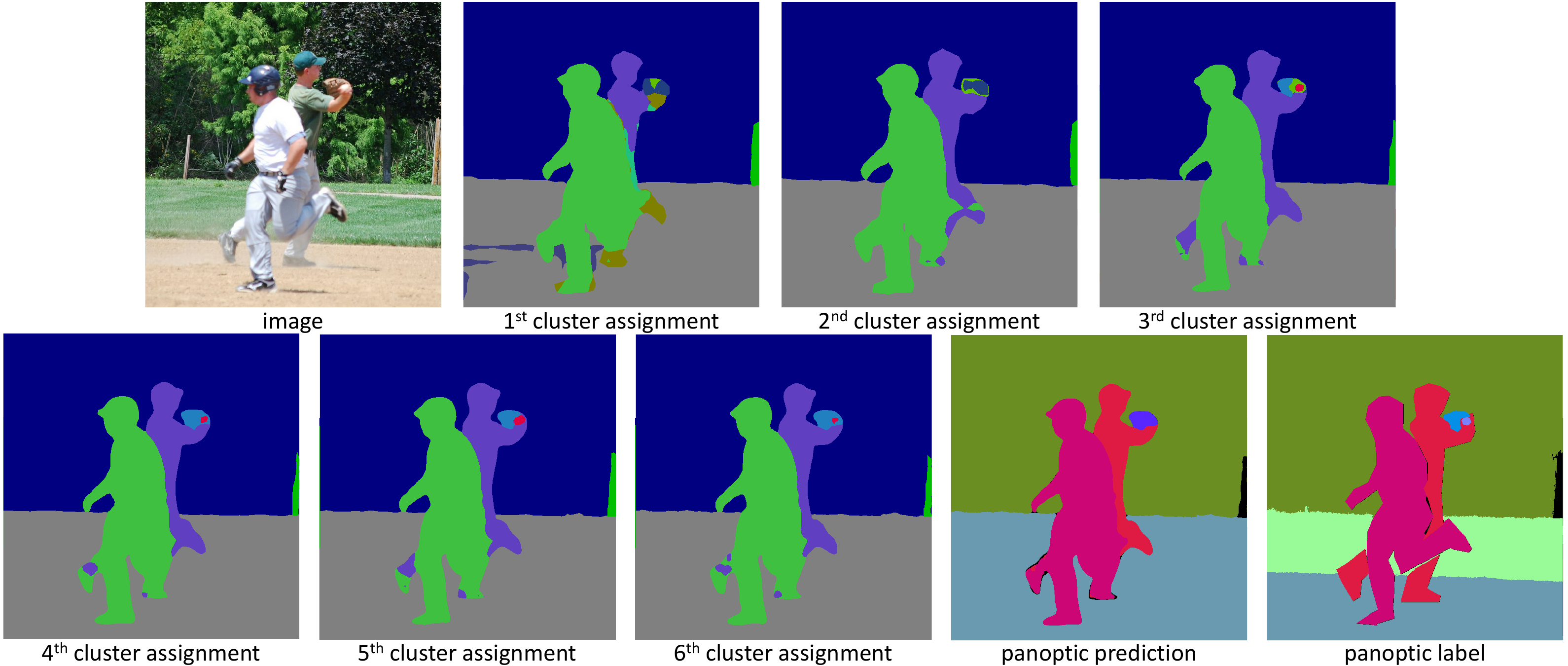}
    \caption{
    \textbf{[Failure mode]} $k$MaX-DeepLab struggles to segment both heavily occluded objects and obscure small objects.
    Specifically, the legs between occluded persons are not well segmented. Additionally, the obscure small baseball is not found at the first two stages. Even though it is recovered in the 3rd clustering stage, it still vanishes in the final prediction. It remains a challenging problem to make the full use of all clustering results to help the final prediction}
    \label{fig:supp_vis_kmax_1}
\end{figure}

\begin{figure}[t]
    \centering
    \includegraphics[width=1.0\linewidth]{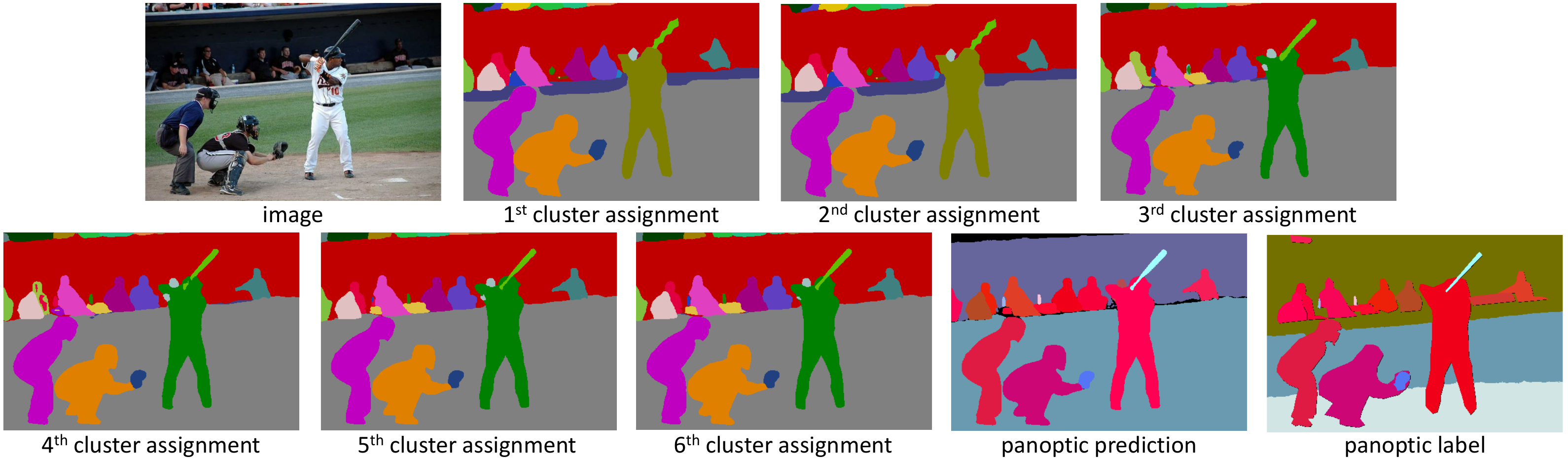}
    \caption{
    \textbf{[Failure mode]} Although $k$MaX-DeepLab shows a strong ability to split images into different regions, it may not yield the correct semantic prediction.
    In this example, $k$MaX-DeepLab is able to segment out the background regions, but fails to predict the correct semantic labels
    }
    \label{fig:supp_vis_kmax_2}
\end{figure}

To better understand the working mechanism behind $k$MaX-DeepLab model, we visualize the $k$MaX-DeepLab clustering process in~\figref{fig:supp_vis_kmax_3} and~\figref{fig:supp_vis_kmax_5}, along with some failure cases in~\figref{fig:supp_vis_kmax_1} and~\figref{fig:supp_vis_kmax_2}.
We utilize $k$MaX-DeepLab with ResNet-50 for all visualizations, including the pixel-cluster assignment (\ie, $\operatornamewithlimits{argmax}_{N}(\mathbf{Q}^{c} \times (\mathbf{K}^{p})^{\mathrm{T}})$ in Eq. (7) of main paper) at each $k$MaX decoder stage and the final panoptic prediction.
In the visualization of pixel-cluster assignments, pixels with the same color are assigned to the same cluster and their features will be aggregated to update the corresponding cluster centers.

As shown in~\figref{fig:supp_vis_kmax_3} and~\figref{fig:supp_vis_kmax_5}, $k$MaX-DeepLab is capable of dealing with small objects and complex scenes, leading to a good panoptic prediction.
We further visualize the failure modes of $k$MaX-DeepLab in \figref{fig:supp_vis_kmax_1} and~\figref{fig:supp_vis_kmax_2}.
$k$MaX-DeepLab has some limitations, when handling heavily occluded objects and predicting correct semantic classes for challenging masks.

\end{document}